\documentclass{preprint}

\usepackage{booktabs}
\usepackage{tabularx}
\usepackage{array}
\usepackage{amsmath,amssymb,amsfonts}
\usepackage{algorithmic}
\usepackage{graphicx}
\usepackage[table]{xcolor}
\usepackage[colorlinks=true]{hyperref}
\usepackage[numbers,comma,sort&compress]{natbib} 

\usepackage{mdframed}
\usepackage[inline]{enumitem}
\usepackage{xspace}

\newcolumntype{C}{>{\centering\arraybackslash}X}
\newcolumntype{R}{>{\raggedleft\arraybackslash}X}

\definecolor{graybg}{HTML}{F5F5F5}
\newcommand\boxedname{Prompt\xspace} 
\newcounter{prompt}
\newenvironment{prompt}[1][]{
    \refstepcounter{prompt}
    \definecolor{graybg}{HTML}{f1f1f1}
    \begin{mdframed}[
        innertopmargin=2pt, 
        innerbottommargin=2pt,
        innerleftmargin=2pt, 
        innerrightmargin=2pt,
        frametitle={\textbf{\boxedname \theprompt:} #1},
        frametitlefont={},
        frametitlerule=true,
        backgroundcolor=graybg]%
    \setlength{\parindent}{0pt}%
    \setlength{\parskip}{1em}%
}
{%
    \par%
    \end{mdframed}%
}

\title{A Source-Grounded Framework for Constructing and Evaluating Progressive Multimodal Diagnostic Dialogues from Clinical Case Reports
}

\author[1,$\dagger$]{Yufan Wang}
\author[1,$\dagger$]{Rui Yang}
\author[2]{Yi Liu}
\author[1]{Yi Lin}
\author[1,*]{Yifan Peng}
\affil[1]{Population Health Sciences,
Weill Cornell Medicine,
New York City, USA}
\affil[2]{Department of Medicine,
Weill Cornell Medicine,
New York City, USA}
\affil[*]{Corresponding author(s). Email(s): \url{yi p4002@med.cornell.edu}}
\affil[$\dagger$]{These authors contributed equally to this work.}

\begin{document}

\maketitle

\begin{abstract}
Clinical diagnosis requires progressive integration of patient history, physical examination, laboratory findings, medical images, and diagnostic-informative tests. However, most multimodal medical benchmarks evaluate fixed inputs or endpoint answers, while fully interactive diagnostic agents conflate evidence selection with evidence interpretation. We present a source-grounded framework to construct progressive multimodal diagnostic dialogues from case reports and an evaluation strategy for assessing MLLMs on final diagnosis, diagnostic reasoning, and image-finding interpretation.  Evaluation on 24 internal medicine case reports showed that our framework can accurately convert case reports into reference dialogues, achieving a diagnosis F1 of 0.99 and a reasoning-quality score of 4.79 out of 5. Evaluation on two frontier MLLMs (o4-mini and Claude Haiku 4.5) achieved reasoning-quality scores of 2.75 and 2.50, respectively, with substantially lower diagnosis, reasoning, and image-finding F1 scores. The results demonstrate that fluent responses do not necessarily reflect evidence-grounded clinical reasoning and highlight the utility of the proposed framework for evaluating multimodal diagnostic reasoning.

\end{abstract}

\begin{keywords}
multimodal large language models \and clinical reasoning \and progressive diagnostic dialogue
\end{keywords}

\section{Introduction}
\label{sec:introduction}

Multimodal large language models (MLLMs) have expanded medical artificial intelligence beyond image classification toward visual question answering, open-ended image interpretation, and diagnostic synthesis. 
In clinical diagnosis, however, information is rarely available as a single fixed input~\cite{lau2018vqarad, liu2021slake, li2023llavamed, moor2023medflamingo}. Clinicians progressively integrate patient history, physical examination, routine laboratory findings, medical images, and diagnosis-informative tests as a case unfolds. Evaluating whether MLLMs can perform this progressive integration is therefore important for assessing their potential role in diagnostic decision support.

Recent work has begun to evaluate medical models through diagnostic
dialogue and interactive consultation~\cite{tu2025conversational, saab2026multimodal, liu2025interactive, sviridov2025_3mdbench}.
However, fully interactive diagnostic agents must simultaneously decide which question or test to request, interpret the evidence returned, and integrate it into an evolving differential diagnosis. Joint evaluation of these capabilities makes it difficult to attribute diagnostic failures to inappropriate test selection, misinterpretation of evidence, or inadequate reasoning; a controlled setting is therefore needed to isolate evidence interpretation and integration before fully assessing interactive diagnostic agents.

Published case reports provide a useful substrate for such controlled
evaluation because they connect patient history, physical examination,
laboratory testing, medical figures, diagnostic reasoning, and final
diagnosis at the patient-case level. PubMed Central (PMC) further
provides machine-readable full-text articles with associated media
files~\cite{nlm2026pmcoa}. However, case reports also contain
information that may directly reveal the answer. Direct conversion of raw articles risks information leakage from figure captions, later article sections, and confirmatory diagnostic results.

Additionally, evaluating MLLM performance in diagnostic dialogues requires more than final-diagnosis accuracy. A model may reach the correct diagnosis
while omitting case-defining evidence, introducing unsupported image
findings, contradicting the source case, or using clinically invalid
intermediate inferences. Conversely, exact matching of complete
responses is too brittle because medically equivalent statements may
differ in wording, specificity, and granularity. These limitations highlight the need for an evaluation approach that assesses not only whether an MLLM reaches the correct diagnosis, but also whether its reasoning is complete, clinically valid, and grounded in the available multimodal evidence.

To bridge these gaps, we present a source-grounded framework for constructing progressive multimodal diagnostic dialogues from case reports, and an evaluation strategy for assessing MLLMs on final diagnosis and diagnostic reasoning. Article-derived references are extracted into computer-interpretable case representations, aligned with local medical images, and separated from the evidence visible during dialogue construction. Reference dialogues and test-model dialogues are generated under the same progressive evidence-disclosure protocol. 
During the evaluation of MLLM in these cases, final diagnoses, diagnostic reasoning, and image findings are assessed using atomic-item alignment. In contrast, overall reasoning quality is assessed using case-level factuality, validity, coherence, and utility (FVCU) scoring. 

The main contributions of this work are threefold:
\begin{enumerate*}[nosep, label=(\roman*)]
\item We introduce a source-grounded framework that converts PMC case reports into computer-interpretable case representations and progressive multimodal diagnostic dialogues.

\item We define a controlled evidence-disclosure protocol that isolates multimodal evidence interpretation and integration from autonomous evidence-selection behavior.

\item We provide an evaluation framework that combines atomic item alignment for final diagnosis, diagnostic reasoning, and image findings with case-level FVCU scoring of clinical reasoning quality.
\end{enumerate*}

\section{Related Work}

Early medical multimodal benchmarks, including VQA-RAD and SLAKE, evaluate responses to predefined questions about medical images. At the same time, LLaVA-Med and Med-Flamingo support more open-ended generative responses to biomedical visual inputs~\cite{lau2018vqarad,liu2021slake,li2023llavamed,moor2023medflamingo}. These resources provide important measurements of multimodal capability but primarily operate under fixed-input settings. More recent systems and benchmarks, including AMIE, multimodal AMIE, MMD-Eval, and 3MDBench, move toward interactive diagnostic consultation and proactive information exchange~\cite{tu2025conversational,saab2026multimodal,liu2025interactive,sviridov2025_3mdbench}. However, fully interactive settings jointly evaluate evidence selection,  interpretation, and diagnostic integration, making it difficult to determine which capability caused a diagnostic failure. Our work instead uses a fixed staged evidence-disclosure protocol to isolate evidence interpretation and integration from autonomous evidence selection.

Open-ended diagnostic dialogue also requires evaluation beyond exact matching or final-answer accuracy. Prior clinical LLM evaluation has used structured judges and checklist-based comparisons between free-form outputs and reference information~\cite{zhang2025llmevalmed}. Such methods can assess whether important clinical content is retained, but component-level alignment alone does not determine whether the complete diagnostic synthesis is clinically sound. We therefore combine atomic-item alignment for final diagnosis, diagnostic reasoning, and image findings with case-level factuality, validity, coherence, and utility scoring adapted from reasoning-trace evaluation~\cite{lee2025evaluating}. This combination localizes specific evidence errors while also assessing the overall quality of the diagnostic reasoning.

\section{Methods}
\label{sec:methods}

\subsection{Study Design and System Overview}
\label{subsec:study-design}

\begin{figure*}[!t]
\centering
\includegraphics[width=0.91\textwidth]{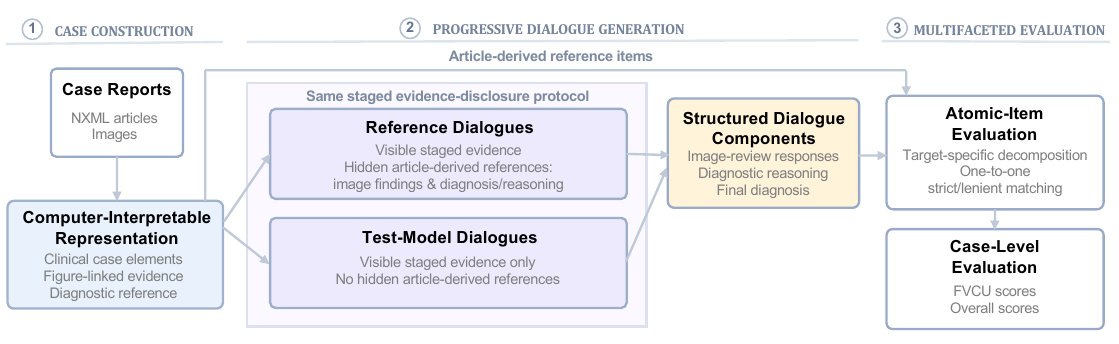}
\caption{Overview of the source-grounded framework for constructing and evaluating progressive multimodal diagnostic dialogues from PMC case reports.}
\label{fig}
\end{figure*}

We developed and evaluated a source-grounded workflow for constructing progressive multimodal diagnostic dialogues from medical case reports.
As summarized in Figure~\ref{fig}, the workflow comprised three stages: computer-interpretable representation extraction, progressive dialogue generation, and multifaceted evaluation.

Throughout this paper, \emph{article-derived references} denote structured, source-supported information extracted from the original case report. These references were organized into a \emph{computer-interpretable case representation} containing clinical case elements, figure-linked evidence, and diagnostic assessment references. \emph{Reference dialogues} denote dialogues generated from the article-derived references. \emph{MLLM-generated dialogues} denote dialogues generated by the evaluated MLLMs. 

\subsection{Input Requirements} 
\label{subsec:case-corpus} 

This framework leverages the JATS-compliant full-text XML format used by PubMed Central (PMC) (\url{https://jats.nlm.nih.gov/nlm-dtd/publishing/}), in which Open Access article packages provide full-text article metadata together with associated media files, including images~\cite{comeau2019pmcbioc}.

A case was eligible if it satisfied all of the following input requirements: 
\begin{enumerate*}[nosep, label=(\roman*)]
\item A parseable XML article. 
\item At least one retrievable image.
\item An extractable article-reported final diagnosis.
\item Sufficient article-reported content to support progressive dialogue construction and evaluation, including patient information, timeline, diagnostic tests (e.g., labs, imaging, pathology), clinical findings, and diagnostic reasoning. 
\end{enumerate*}

For cases containing multiple figures, we preserved the image order defined in the original article. Figure references were first identified in the article source and then matched to the corresponding image files. This ordering preserved the source publication progression of visual evidence and ensured that multi-image cases remained aligned with the accompanying textual narrative and diagnostic reasoning.

\subsection{Computer-interpretable representation extraction}
\label{subsec:reference-construction}

Each case report was processed using a source-grounded LLM extractor to construct a computer-interpretable representation. The extraction schema was informed by clinically relevant items of the CARE case-report checklist, including patient information, clinical findings, timeline, diagnostic assessment, interventions or outcomes when reported, and diagnostic rationale~\cite{gagnier2014care}.

The extractor was instructed to return source-supported text spans or minimally reorganized source content when available, and to leave fields empty when the corresponding information was not explicitly reported (Prompt~\ref{prompt:extraction}). The structured output mapped case report information into three workflow-specific groups: clinical case elements, figure-linked evidence, and diagnostic assessment references.

\begin{prompt}[Source-grounded extraction]
\label{prompt:extraction}

Extract only information explicitly stated in the source case report. Every non-empty field must be directly supported by the source text; otherwise, leave the field empty. Do not infer, speculate, normalize, or generate new medical content.

Organize the output into clinical case elements, figure-linked evidence, and diagnostic assessment references. Clinical case elements must exclude the final diagnosis and image-derived diagnostic information. Keep routine laboratory findings separate from key diagnosis-informative findings. For figure-linked evidence, extract only source-text descriptions of the figure identifier, image type or modality, image finding, and diagnostic significance; do not infer findings from image pixels. For diagnostic assessment references, extract the article-reported final diagnosis and source-supported diagnostic reasoning.
\end{prompt}

\subsubsection{Clinical case elements} captured patient information, primary concern, medical, family, and psychosocial history, physical examination, and important clinical findings, routine laboratory testing, key diagnosis-informative findings, diagnostic methods, and timeline from the episode of care. Routine laboratory testing captured baseline clinical tests, whereas key diagnosis-informative findings captured special tests or results that directly supported the diagnosis, such as immunohistochemistry, molecular or genetic testing, microbiological identification, tumor markers, or flow cytometry. The diagnostic tests field contained only the names or types of procedures and examinations performed, with all associated diagnostic results excluded.

\subsubsection{Figure-linked evidence} included the image type or modality, clinical findings from the image, and diagnostic significance when available. Diagnostic significance was defined as source-article text that explicitly stated how an image finding supported, confirmed, suggested, or ruled out a diagnosis or differential diagnosis. These fields were extracted from article text, figure captions, or adjacent figure legends when explicitly available. The extraction process did not inspect or interpret image pixels.

\subsubsection{Diagnostic assessment references} included the article-reported final diagnosis and source-supported sentences describing the
case-specific evidence or diagnostic process leading to that diagnosis.
The extractor was prohibited from rewriting the diagnosis or generating
additional clinical reasoning.

\subsection{Progressive Dialogue Construction}
\label{subsec:dialogue-generation}

\begin{figure}[t] 
\centering 
\includegraphics[width=0.5\columnwidth]{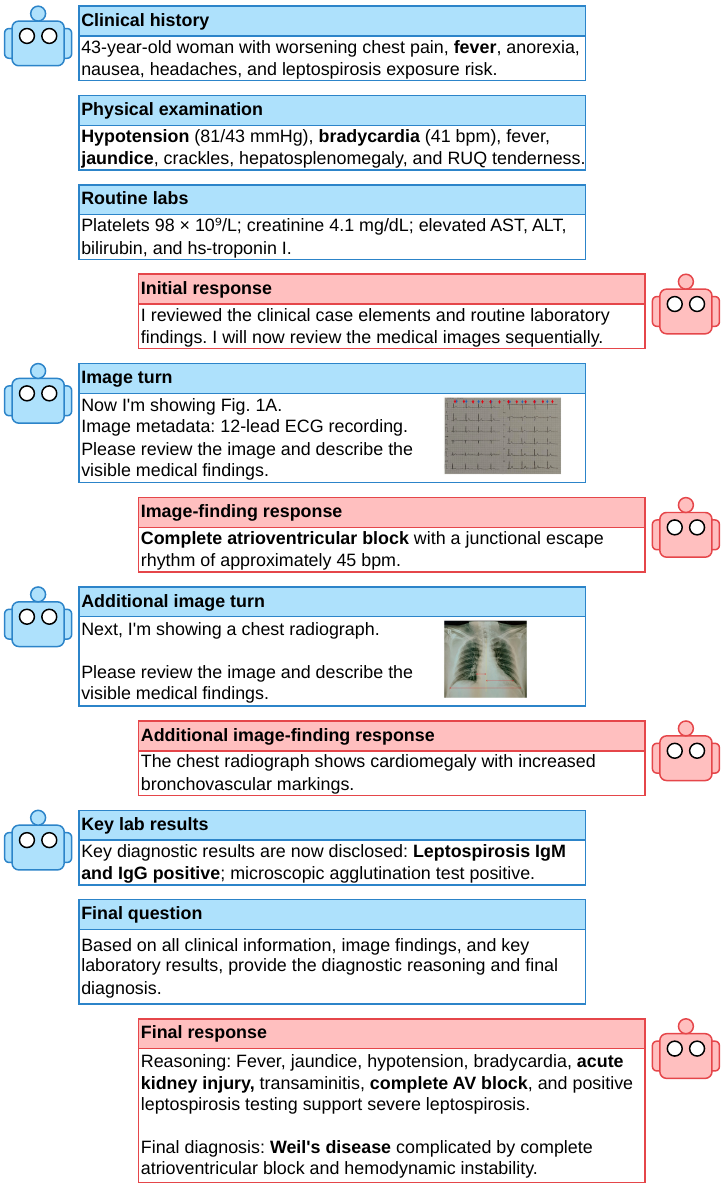} 
\caption{Example of a progressive multimodal diagnostic dialogue
constructed from a PMC case report under the staged evidence-disclosure protocol.}
\label{fig:dialogue-protocol} 
\end{figure}

Dialogues were constructed as progressive multimodal diagnostic dialogues using the computer-interpretable case representations described above. Figure~\ref{fig:dialogue-protocol} summarizes the saved dialogue QA format.

The progressive evidence-disclosure protocol followed a clinical workflow structure inspired by prior work on autonomous medical AI agents, in which disease history, physical examination, laboratory testing, imaging, and downstream diagnostic actions are represented as sequential clinical information nodes~\cite{ferber2026autonomous}. In our implementation, the model first received the clinical history, including patient demographics, chief complaint or primary concern, and, when available, the history of present illness. It then received physical examination findings and routine laboratory results. Key diagnosis-informative findings were withheld from the initial clinical turn because highly diagnostic laboratory, molecular, microbiologic, genetic, pathology, or immunohistochemistry results could directly reveal the final diagnosis before image review.

Medical images were then presented sequentially in source-article order. Each image-review turn included the raw image and non-diagnostic metadata, such as examination type or image modality. The prompt instructed the evaluated LLM to describe visible clinical findings, distinguish direct observations from interpretation, and use cautious diagnostic language when uncertain. Reference image-review responses were constructed from the article-derived image finding and diagnostic significance for each figure; the generator was instructed to preserve the medical meaning of these references while rewriting them as concise physician-style responses without unsupported additional findings.

After all image-review turns were completed, key diagnosis-informative findings were disclosed, and the final synthesis turn was constructed. This turn required diagnostic reasoning and a final diagnosis based on all clinical information, image review responses, and key laboratory or diagnostic findings. For the reference dialogue, the final diagnostic reasoning was generated using the article-derived diagnostic reasoning.

\subsection{Atomic Item Evaluation}
\label{subsec:automated-evaluation}

Generated dialogues were evaluated using an item-decomposed
LLM-as-judge workflow for final diagnosis, diagnostic reasoning, and
image findings. For each target, the article-derived reference and dialogue output were decomposed into target-specific atomic items, and
a separate LLM matcher compared the resulting items under one-to-one
matching constraints (Prompt~\ref{prompt:item-matching}).

From the resulting atomic item matches, true positives (TPs) were defined as generated items matched to reference items under the specified criterion. False negatives (FNs) were reference items without a corresponding generated item, representing missing evidence. False positives (FPs) were generated items without a corresponding reference item, representing extra generated content. For strict metrics, only strict matches were counted as TPs; for lenient metrics, both strict and lenient matches were counted as TPs. 

\begin{prompt}[Atomic item matching]
\label{prompt:item-matching}
Match article-derived reference items to dialogue-output items. Use extracted items only, not raw source paragraphs. Do not further decompose items, calculate metrics, or match by list order.

Apply one-to-one matching: each item may be used at most once. Return strict matches, lenient matches, and contradictions. \textbf{Strict matches} preserve the clinically essential meaning of the reference item. \textbf{Lenient matches} share the same diagnostic or visual evidence core but may be broader, narrower, incomplete, or less precise. \textbf{Contradictions} require direct clinical incompatibility. Vague overlap, same-organ overlap, broad disease-category overlap, or matches requiring external inference should be treated as no match. Unmatched dialogue-output items should not automatically be classified as contradictions.
\end{prompt}

\subsubsection{Final-Diagnosis Evaluation}

Atomic items included diseases, syndromes, etiologies, causal organisms, disease subtypes, named complications, and causal diagnoses asserted as final diagnoses. The item decomposer preserved clinically essential modifiers when they were part of the diagnosis, including etiology, causal organism, anatomical site, disease subtype, named complication, severity when clinically diagnostic, and causal relationships such as ``caused by,'' ``secondary to,'' or ``associated with.'' We report strict precision, recall, and F1 because omission of these modifiers can change the clinical meaning of the diagnosis.

\subsubsection{Diagnostic-Reasoning Evaluation}

Atomic items were clinically meaningful steps used to support, exclude, differentiate, localize, characterize, or explain the diagnosis. The same decomposition scheme was applied to the article-derived diagnostic reasoning. Still, the LLM was instructed to keep diagnostically relevant claims made by the evaluated model, even when they were incorrect, unsupported, or inconsistent with the reference diagnosis. 
This design ensures that missing evidence, unsupported additions, and contradictory reasoning are identified during the matching stage rather than being filtered out during decomposition.
We report lenient precision, recall, and F1 because clinically relevant reasoning can differ in wording or granularity while preserving the same diagnostic evidence. 

\subsubsection{Image-Finding Evaluation}

An image-finding item was defined as an atomic, clinically meaningful visual observation extracted from either an article-derived image finding or an image-review response. Image-finding items could represent anatomy, lesion morphology,
location, laterality, distribution, density, signal, enhancement,
color, staining pattern, cellular morphology, nuclear features, or
tissue architecture. For example, in one pathology figure, the visible-image units included ``spindle-shaped cells,'' ``fasciculated and disarrayed architecture,'' and ``nuclei in a palisading pattern.'' In contrast, diagnostic impressions such as schwannoma or nerve-sheath tumor were not treated as visible-image units.  Image-finding alignment was computed only for figures with both an article-derived image finding and an available generated image-review response. We report lenient recall and F1 because visible findings may be described at different levels of specificity.

\subsection{Case-Level FVCU Scoring}

Diagnostic reasoning quality was further evaluated at the case level using a clinical adaptation of reasoning-trace evaluation criteria inspired by prior work~\cite{lee2025evaluating} (Prompt~\ref{prompt:fvcu}). 

For each case, the FVCU judge received a structured summary of the item-level evaluation, containing the reference diagnosis and diagnostic evidence, the model-generated diagnosis and reasoning, and the item-level alignment results, including matched, missing, extra, and contradictory evidence. Based on this information, the judge assigned a score from 1 to 5 for each FVCU dimension.

\textbf{Factuality} assessed whether case-specific clinical, imaging, laboratory, pathology, microbiology, molecular, or genetic claims were supported by the available evidence. \textbf{Validity} assessed whether the diagnostic inferences were clinically reasonable given the evidence presented in the model's reasoning. \textbf{Coherence} assessed internal consistency and logical organization. \textbf{Utility} assessed whether the reasoning covered, prioritized, and synthesized the case-defining evidence needed to support the article-reported diagnostic pathway. 

\begin{prompt}[Case-level FVCU judging]
\label{prompt:fvcu}
Evaluate the model-generated diagnostic reasoning using the structured
item-level evaluation summary, including the reference diagnosis and
evidence, model-generated diagnosis and reasoning, and matched, missing,
extra, or contradictory evidence.
Assign a score from 1 to 5 for each predefined FVCU dimension and provide
a brief evidence-grounded justification. Then assign a separate overall
reasoning-quality score from 1 to 5 as a holistic judgment rather than
the arithmetic mean of the four dimension scores. Major factuality or
validity errors, or omission of case-defining evidence, should lower the
overall score.
\end{prompt}

\section{Experiments}
\label{sec:experimental-design}

\subsection{Evaluation Dataset}
\label{subsec:experimental-settings}

The experimental corpus consisted of 24 publicly available internal medicine case reports from PMC. Across these cases, the image set covered diverse clinical modalities, including radiology and angiography, pathology and cytology, endoscopy, ophthalmic imaging, echocardiography, and electrocardiography. Each case contained a median of 3.5 images (range: 1--8).

\subsection{Framework Implementation and Evaluation}
\label{subsec:framework-implementation}

\subsubsection{Setup}

Within the framework, GPT-4.1 was used for source-grounded information extraction, reference dialogue generation, and item decomposition for automated evaluation. GPT-5.1 was used for semantic item matching and case-level FVCU scoring.

\subsection{Benchmarking Existing MLLMs}
We benchmarked two MLLMs, o4-mini \cite{openai2025o4mini} and Claude Haiku 4.5
\cite{anthropic2025haiku45}. Both followed identical staged
prompts, source-image order, and evidence-disclosure schedules.
Initially, each received only clinical case elements and routine
laboratory findings. At each image-review turn, the model received
the raw image and non-diagnostic metadata, including the figure
identifier and modality when available. Key diagnosis-informative
findings were disclosed only during final synthesis. Neither model
received figure captions, article-derived image findings, diagnostic
reasoning references, or the final diagnosis.

Final-diagnosis and reasoning metrics were macro-averaged over
24 cases. Image-finding metrics were macro-averaged over 86
evaluable image turns, defined as turns with both an article-derived
finding and a generated response. FVCU scores were averaged over
24 cases.

\section{Results and Discussion}
\label{sec:results}

\subsection{Reliability of Reference Dialogue Construction}
\label{subsec:results-reference}

\begin{table}[t]
\centering

\caption{Evidence alignment and FVCU scores across dialogue settings. Alignment values are macro-averaged; FVCU values are mean 1--5 scores.}
\label{tab:combined_scores}
\footnotesize
\setlength{\tabcolsep}{5pt}
\renewcommand{\arraystretch}{1.08}
\begin{tabularx}{.7\textwidth}{lCCC}
\toprule
\textbf{Metric} &
\textbf{Reference Dialogue} &
\textbf{o4-mini} &
\textbf{Claude} \\
\midrule
\rowcolor{gray!20}Final Diagnosis & & & \\
~~~~Strict Precision & 0.99 & 0.32 & 0.20  \\
~~~~Strict Recall & 1.00 & 0.45 & 0.39 \\
~~~~Strict F1     & 0.99 & 0.35 & 0.25 \\
\rowcolor{gray!20}Diagnostic Reasoning & & & \\
~~~~Lenient Precision & 0.67 & 0.25 & 0.19 \\
~~~~Lenient Recall & 0.77 & 0.34 & 0.38 \\
~~~~Lenient F1     & 0.69 & 0.28 & 0.24 \\
\rowcolor{gray!20}Image Finding & & & \\
~~~~Lenient Precision & 0.62 & 0.12 & 0.06 \\
~~~~Lenient Recall & 0.81 & 0.41 & 0.39 \\
~~~~Lenient F1     & 0.67 & 0.17 & 0.11 \\
\rowcolor{gray!20}FVCU & & & \\
~~~~Factuality & 4.62 & 3.00 & 2.88 \\
~~~~Validity   & 4.92 & 3.12 & 2.62 \\
~~~~Coherence  & 5.00 & 4.54 & 4.29 \\
~~~~Utility    & 4.25 & 2.29 & 2.12 \\
\rowcolor{gray!20}Overall    & 4.79 & 2.75 & 2.50 \\
\bottomrule
\end{tabularx}
\end{table}

We first evaluated whether reference dialogues preserved article-derived diagnostic information after conversion into progressive dialogue format. For each target, reference-dialogue outputs were aligned with the corresponding article-derived reference items. Atomic item evaluation showed high preservation of the article-reported final diagnosis, with macro-averaged strict precision, recall, and F1 of 0.99, 1.00, and 0.99, respectively (Table~\ref{tab:combined_scores}).

Reference dialogues also preserved most article-derived reasoning and image-finding evidence, although precision was lower than recall for both components. 
Macro-averaged lenient F1 scores were 0.69 for reasoning and 0.67 for image
finding. These results reflected core evidence preservation with additional or
differently segmented items introduced during concise dialogue
reformulation. The high overall FVCU score of 4.79 further supported their case-level reliability.

\subsection{Performance of Existing Multimodal LLMs}
\label{subsec:results-alignment}

Both test-model conditions showed substantially lower alignment
with the article-derived references (Table~\ref{tab:combined_scores}). Final-diagnosis strict F1 scores were 0.35 for o4-mini and 0.25 for Claude, with strict precision of only 0.32 and 0.20, respectively.

The largest performance gaps were observed for diagnostic reasoning and image finding. Lenient reasoning F1 was 0.28 for o4-mini and 0.24 for Claude. Lenient image-finding F1 was 0.17 for o4-mini and 0.11 for Claude. The low image-finding precision of the test models, 0.12 for o4-mini and 0.06 for Claude, indicates that many generated visual claims were not supported by the article-derived figure evidence.

Both models also produced relatively coherent reasoning narratives, with coherence scores of 4.54 for o4-mini and 4.29 for Claude. However, their factuality, validity, utility, and overall reasoning-quality scores were lower than those of the reference dialogues. Utility showed the largest gap, decreasing to 2.29 for o4-mini and 2.12 for Claude. This pattern suggests that fluent organization alone was insufficient for clinically useful diagnostic reasoning; evidence coverage and factual grounding were the main limiting factors.

\subsection{Error Analysis}
\label{subsec:results-error}

For the reference dialogue, lower reasoning and image-finding F1 scores were mainly driven by unmatched generated items introduced during dialogue reformulation. The item-level extra generated-item rate, defined as $\frac{FP}{TP+FP}$, was 0.03 for final diagnosis, 0.35 for diagnostic reasoning, and 0.38 for image findings. These unmatched items were not necessarily clinically implausible; rather, they often reflected additional explanatory wording or differences in item granularity relative to the article-derived reference.

A separate source-side mismatch arose in one reference image turn. The article-derived ECG finding described ``total atrioventricular block with a junctional escape rhythm of 45 bpm.'' The generated dialogue preserved the core findings of atrioventricular block and junctional escape rhythm. However, the source-side extracted image-finding items also included annotation details, such as color markers for P waves and QRS complexes, that were not repeated in the concise dialogue response and therefore contributed to unmatched reference items.

The evaluated MLLMs showed a different error pattern. Their image-finding extra-generated-item rates were much higher, 0.89 for o4-mini and 0.94 for Claude, indicating that their low image-finding F1 reflected not only missed article-derived findings but also many generated visual claims that were not supported by the article-derived figure evidence.

\subsection{Limitations}

Several limitations should be considered. First, the study used a fixed, progressive evidence-disclosure protocol with predefined prompts, so it evaluated the interpretation and integration of provided evidence rather than autonomous follow-up questioning, test selection, evidence prioritization, or stopping behavior. Second, the corpus was limited to 24 publicly available internal medicine case reports and two evaluated MLLMs; published case reports may overrepresent unusual or diagnostically challenging presentations and may not reflect routine clinical records, common disease distributions, or other specialties. Third, although the article-derived references were source-grounded, they were not independent or exhaustive expert annotations of image pixels or diagnostic reasoning. Unmatched generated items may therefore represent unsupported claims, wording, or differences in granularity, or clinically plausible findings not reported in the source article. Fourth, reference dialogue generation, item decomposition, semantic matching, and FVCU scoring relied partly on LLM-based processing, introducing potential judge-model bias and sensitivity to prompt design. Because case-level FVCU scoring was conditioned on the preceding item-decomposition and semantic-matching outputs, it did not provide an
independent validation of these evaluation stages. Moreover, the adapted FVCU rubric has not yet been validated as a clinical rating instrument, and formal clinician scoring was not included in the present study. Preliminary clinician feedback suggested that a single holistic score was difficult to apply consistently to reconstructed QA-format
dialogues, which may not fully preserve the sequential diagnostic narrowing represented in the original case reports. Future work will therefore develop dimension-specific clinician rating criteria and directly compare clinician assessments with LLM-as-judge scores.

\section{Conclusion}

We introduced a source-grounded framework for evaluating evidence
integration in progressive multimodal diagnostic dialogues constructed
from PMC case reports. Under a shared staged evidence-disclosure
protocol, reference dialogues largely preserved article-derived
diagnoses and evidence, whereas two evaluated MLLMs showed substantially
lower alignment, especially for image findings. Their high coherence
but lower factuality and utility demonstrate that fluent clinical
reasoning is not necessarily evidence-grounded. This framework provides
a controlled basis for evaluating multimodal evidence interpretation
and developing adaptive diagnostic agents and clinician-facing
evaluation protocols.

\section*{Acknowledgment}

This work was supported by the National Library of Medicine grants R01LM014344 and R01LM014573.

\bibliographystyle{unsrtnat}
\bibliography{ref}

\end{document}